\documentclass[letterpaper]{article} 
\usepackage[preprint]{aaai2027}  
\usepackage[hyphens]{url}  
\usepackage{graphicx} 
\usepackage{natbib}  
\usepackage{caption} 
\usepackage{algorithm}
\usepackage{algorithmic}
\usepackage{amsmath}
\usepackage{amssymb}
\usepackage{booktabs}
\usepackage{multirow}
\usepackage{bbding}
\usepackage{makecell}
\usepackage{bm}
\usepackage{xcolor}
\usepackage{colortbl}
\usepackage{newfloat}
\usepackage{listings}
\DeclareCaptionStyle{ruled}{labelfont=normalfont,labelsep=colon,strut=off} 
\floatstyle{ruled}
\newfloat{listing}{tb}{lst}{}
\floatname{listing}{Listing}

\usepackage{booktabs}

\title{Focus and Supplement: Dual-Enhanced Vision Transformer for Multi-Class Anomaly Classification}
\author{
    Xurui Li, Enjie Xu, Chenzhou Li, Shilei Zeng, Dayou Huang, Tianyi Ma, Yu Zhou\corresponding
}
\affiliations{
    School of Electronic Information and Communications, Huazhong University of Science and Technology \\
    \{xrli\_plus, xej, chenzhouli, shlzeng, d202581412, D202481253, yuzhou\}@hust.edu.cn
}

\begin{document}

\maketitle

\begin{abstract}
Multi-class anomaly classification in industrial vision remains challenging due to noisy/incomplete anomaly representations and the unknown number of anomaly classes.
To overcome this, we propose MACO, a novel multi-class anomaly classification framework that learns comprehensive representations and dynamically estimates class number without prior knowledge.
First, a soft-focus attention uses anomaly maps to concentrate on relevant abnormal regions, while suppressing background noise.
Second, auxiliary classification (\texttt{[A-CLS]}) tokens complement the \texttt{[CLS]} token.
They collectively attend to diverse anomaly sub-regions, yielding more holistic and discriminative features.
These \texttt{[A-CLS]} tokens are also effective across more tasks and domains.
To infer the class number, we propose Correlation-based Number Estimation strategy.
It computes the average correlation among labeled classes and transfers its separability cue to the unlabeled set.
Experiments on MVTec AD and MTD datasets demonstrate our superiority.
Under known class number, MACO improves ARI by 6.5\% and \textbf{16.3\%} on both datasets, respectively.
In the more challenging unknown number scenario, it achieves an \textbf{11.2\%} NMI gain on MTD and outperforms existing number estimation strategies by \textbf{24.1\%} UPS on MVTec AD.
Code will be released at https://github.com/HUST-SLOW/MACO.
\end{abstract}

\section{Introduction}
\label{sec:intro}
In industrial visual inspection, anomaly detection \cite{eccv2024glass, CPR, ICLR2024MuSc} has achieved significant progress in localizing anomaly regions.
However, simply identifying the presence of an anomaly is often insufficient for practical applications, such as root cause analysis and process optimization.
Therefore, it is equally critical to recognize fine-grained anomaly classes.
This task, known as multi-class anomaly classification, becomes particularly challenging when dealing with unknown anomaly types in real-world production environments.

\begin{figure}[t]
\begin{center}
\includegraphics[width=0.48\textwidth]{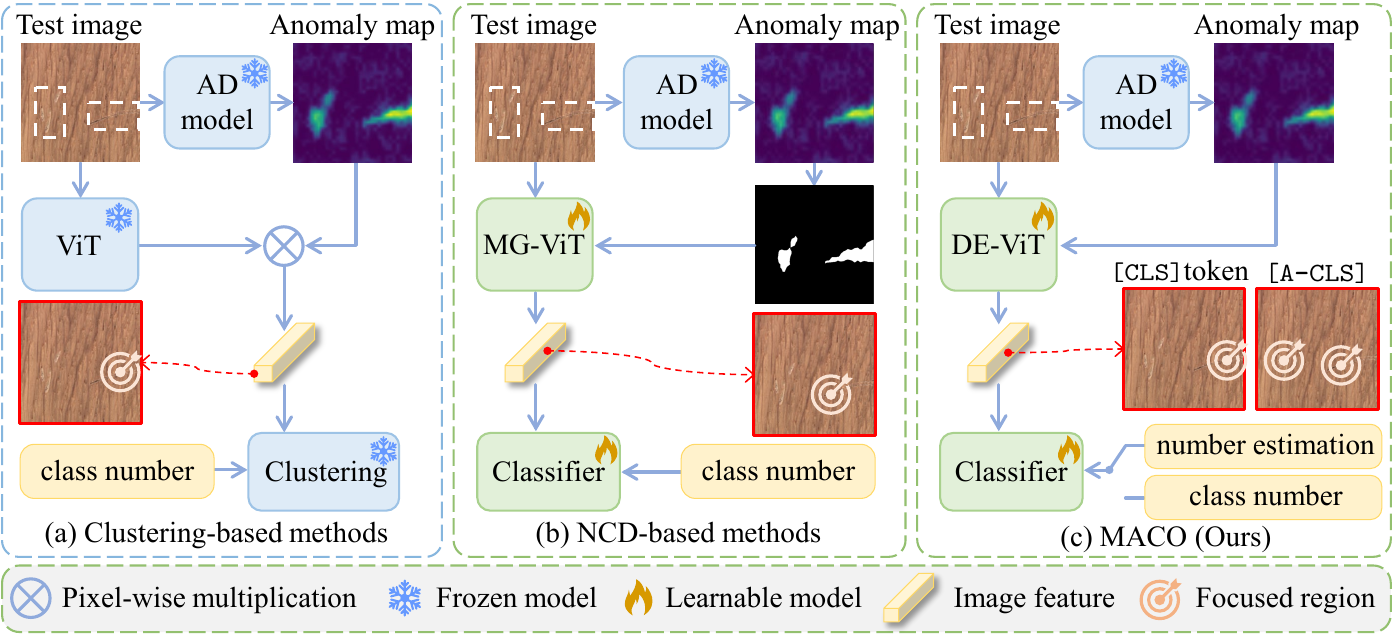}
\caption{
(a) Clustering‑based methods focus on the highest‑probability region.
(b) NCD‑based methods capture only a partial region.
(c) Our MACO attends to diverse abnormal sub‑regions, yielding a complete feature representation.
}
\vspace{-1.5\baselineskip}
\label{fig:intro}
\end{center}
\end{figure}

Current multi-class anomaly classification methods could be divided into clustering-based (Fig.~\ref{fig:intro}(a)) and Novel Class Discovery (NCD)-based (b) approaches (Fig.~\ref{fig:intro}(b)), which face three common limitations.
First, \textbf{noisy representations}: existing methods are susceptible to background interference or noisy anomaly maps, leading to contaminated features that reduce discriminability.
Second, \textbf{incomplete representations}: a single \texttt{[CLS]} token in ViT often attends to only a local part of the anomaly region, failing to capture its full semantic diversity.
Third, \textbf{inability to handle unknown class number}: current methods require the number of anomaly classes to be predefined, which limits their applicability in real industrial scenarios where this number is often unknown.

To address these issues, we propose MACO, an NCD-based framework that learns complete anomaly representations and estimates the number of classes (Fig. \ref{fig:intro}(c)).
We first propose a Dual-Enhanced Vision Transformer (DE-ViT) to extract an anomaly feature for each image.
It incorporates a Soft-focused Attention that operates on anomaly maps.
This design allows the model to focus on anomalies, even those with lower responses, thereby reducing background noise and mitigating noisy representations.
To overcome the incomplete anomaly representation of a single \texttt{[CLS]} token, we design Auxiliary Classification (\texttt{[A-CLS]}) Tokens that collectively attend to diverse abnormal sub-regions, yielding a more holistic and discriminative representation.
These \texttt{[A-CLS]} tokens also prove effective across a wider range of domains and tasks.
To infer the class number, we propose a Correlation-based Number Estimation strategy.
It estimates the class number by computing the average correlation among labeled classes, and transferring this separability cue to the unlabeled set.
We further design an Under-estimation Penalized Score (UPS) tailored for evaluating the estimation accuracy.
Experiments show that our MACO significantly outperforms existing methods on MVTec AD and MTD datasets.
With a known class number, it improves ARI by 6.5\% and \textbf{16.3\%}, respectively.
Under the more challenging unknown number setting, it achieves an \textbf{11.2\%} NMI gain on MTD.
For class number estimation, it improves UPS by \textbf{24.1\%} and \textbf{40\%} on both datasets.

Our contributions are summarised as follows:
\begin{itemize}
\item We propose the Dual-Enhanced Vision Transformer (DE-ViT), which leverages a soft-focus attention and auxiliary classification tokens to extract discriminative and comprehensive anomaly representations.
\item To the best of our knowledge, this is the first multi-class anomaly classification method capable of operating without prior knowledge of the class number.
The proposed Correlation-based Number Estimation strategy is the first to use labeled-class prototypes to estimate inter-prototype gaps among unlabeled classes, thereby accurately determining the number of unlabeled classes.
\item Our approach achieves significant improvements under both known and unknown class number settings on MVTec AD and MTD, with ARI gains up to 16.3\% and NMI gains of 11.2\%, respectively.

\end{itemize}

\begin{figure*}[t]
\begin{center}
\includegraphics[width=1.0\textwidth]{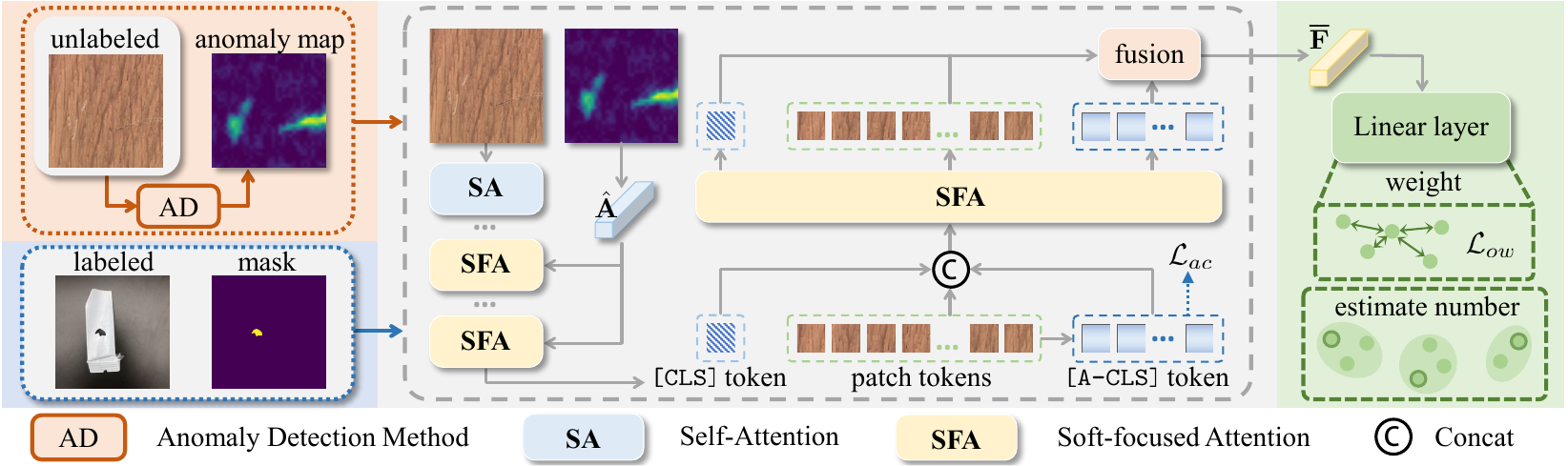}
\caption{
Our pipeline. Anomaly features are extracted by our proposed DE-ViT, which incorporates Soft-focused Attention and Auxiliary Classification Tokens.
These features are classified using a linear classifier regularized by an Orthogonal Weight Loss.
A Correlation-based Number Estimation strategy dynamically estimates the unknown class number if it is unavailable.}
\label{fig:pipeline}
\end{center}
\end{figure*}

\section{Related Work}
\label{sec:related}
\subsection{Multi-class Anomaly Classification}
\subsubsection{Clustering-based methods}
Existing clustering-based approaches for multi-class anomaly classification commonly rely on pre-trained feature extractors, which makes them struggle to capture fine-grained anomalies.
To address this, AC \cite{WACV2023AC} introduces an anomaly map-weighted feature aggregation to focus on regions likely to be anomalies.
Similarly, Uniformaly \cite{pr2026uniformaly} selects the patch feature with high anomaly scores to improve cluster accuracy.

\subsubsection{NCD-based methods}
Novel Class Discovery (NCD) is initially introduced in \cite{ICCV2019DTC}, with early approaches adopting a two-stage pipeline \cite{ICCV2019DTC, iclr2018transfor}.
Recent methods have shifted to one-stage frameworks \cite{TPAMI2021autonovel, NIPS2023uGCD} that jointly use labeled and unlabeled data, reducing bias toward known classes.
However, these methods focus on natural scenes and are hard to handle subtle anomalies.
In industrial scenes, AnomalyNCD \cite{cvpr2025anomalyncd} employs a mask-guided attention to focus on anomaly regions.
Nonetheless, this strategy requires binarizing anomaly maps.
Any anomalies missed during binarization step will inevitably lead to misclassification.
To address this limitation, we propose a Soft-focused Attention that maintains attention to potential anomalies even under incomplete or noisy anomaly maps.

\subsection{\texttt{[CLS]} token of vision transformer}
The \texttt{[CLS]} token in Vision Transformers serves as a global image representations for classification, aggregating patch tokens information via self-attention.
Methods such as the DINO family of models \cite{iccv2021dino, mlrj2024dinov2, arxiv2025dinov3} use it to highlight salient regions, while multimodal CLIP-based models \cite{ICML2021CLIP, nips2025gcl, icml2025Llip} align it with text embeddings via contrastive learning.
However, a single \texttt{[CLS]} token only attends to a local region and brings incomplete anomaly representation,
particularly in industrial anomaly classification, where it tends to overfit to dominant features and ignore subtle inter-anomaly distinctions.
Although \texttt{[reg]} tokens \cite{ICLR2024vit-reg} improve feature smoothness by absorbing redundant information,
they are excluded from the final output and lack explicit mechanisms to prioritize task-relevant regions.
To overcome these issues, we introduce Auxiliary Classification Tokens, which are retained in the output and supplement the \texttt{[CLS]} token by anomaly-specific cues to ensure comprehensive anomaly representation. 

\subsection{Estimating the class number}
In industrial scenarios, the number of anomaly classes is often unknown.
To estimate the number of unlabeled classes, DTC \cite{ICCV2019DTC} selects the class number that maximizes clustering metrics across epochs, while DeepDPM \cite{cvpr2022deepdpm} trains a clustering network that iteratively splits and merges clusters.
PromptCCD \cite{iccv2023promptccd} initializes a candidate number and refines it via splitting decisions.
ProtoGCD \cite{pami2025protogcd} evaluates different numbers using a proposed protoscore.
DAEM \cite{tnnls2025daem} combines the Elbow method and Silhouette score \cite{CAM1987silhouettes}, whereas GCD \cite{CVPR2022GCD} and CMS \cite{cvpr2024cms} determine the number by maximizing accuracy on labeled classes.
These methods either require training additional networks or repeatedly clustering all samples, which becomes computationally prohibitive with large-scale data.
To address this, we propose a Correlation-based Number Estimation strategy.
It leverages prototypes of labeled classes to estimate the inter-prototype gaps among unlabeled classes.

\section{Preliminary}
\label{sec:preliminary}
\subsubsection{Problem Definition}
Following the AnomalyNCD \cite{cvpr2025anomalyncd} framework, we address the problem of novel anomaly class discovery using two datasets:
unlabeled images $\mathcal{D}^\mathbf{u}=\{ I_i^\mathbf{u} | i\in[1,N^\mathbf{u}]\}$ containing novel anomalies and labeled anomaly images $\mathcal{D}^\mathbf{l}=\{(I_i^\mathbf{l}, y_i^\mathbf{l})|i\in[1,N^\mathbf{l}]\}$ with $\mathcal{C}^\mathbf{l}$ known classes, where $y_i^\mathbf{l}$ is the one-hot class label.
The objective is to categorize $\mathcal{D}^\mathbf{u}$ into $\mathcal{C}^\mathbf{u}$ novel classes. In our approach, we consider two distinct experimental settings: one where $\mathcal{C}^\mathbf{u}$ is known a priori, and another more challenging setting where $\mathcal{C}^\mathbf{u}$ is unknown.

\subsubsection{Novel Class Discovery}
Existing NCD methods \cite{MICCAI2023MedNCD, ncdssCVPR} typically use a contrastive learning framework to adaptively learn from the unlabeled set $\mathcal{D}^\mathbf{u}$ while integrating knowledge from a labeled set $\mathcal{D}^\mathbf{l}$.
The framework commonly adopts a teacher-student architecture, where both networks share a common ViT \cite{ICLR2021ViT} backbone $f(\cdot)$,
but are equipped with separate classification heads $\mathcal{H}(\cdot)$ that operate under different softmax temperatures.
For each input image $I_{i}$, two distinct views $\tilde{I}_{i}$ and $\hat{I}_{i}$ are generated via data augmentation.
These views are then fed into the teacher and student networks, respectively.
The student one is trained to classify all images,
whereas the teacher one generates pseudo-labels for unlabeled images.
Specifically, the teacher network computes logits
$\hat{l}_{i}\!=\!\mathcal{H}(f(\hat{I}_{i}))$
and $\tilde{l}_{i}\!=\!\mathcal{H}(f(\tilde{I}_{i}))$ in $\mathbb{R}^{\mathcal{C}^\mathbf{l}+\mathcal{C}^\mathbf{u}}$.
These logits are subsequently converted into sharp pseudo-labels $\hat{q}_{i},\tilde{q}_{i}$ 	
 using a softmax function with a small temperature $\tau_t$.

The optimization objective integrates representation learning and classification losses.
Specifically, $\mathcal{L}_{\text{con}}^s$ \cite{NIPS2020supervised} and $\mathcal{L}_{\text{con}}^u$ \cite{ICML2020SimCLR} denote supervised and self-supervised contrastive losses applied to labeled and unlabeled images, respectively.
Classification employs $\mathcal{L}_{\text{cls}}^\mathbf{l}$ for ground-truth supervision and $\mathcal{L}_{\text{cls}}^\mathbf{u}$ for pseudo-label supervision, regularized by mean-entropy maximization $\mathcal{L}_{\text{reg}}$ \cite{NIPS2020entropy1}.
The composite loss function is:
\begin{equation}
\mathcal{L}_\text{NCD} = \lambda(\mathcal{L}_{\text{con}}^s+\mathcal{L}_{\text{cls}}^\mathbf{l}) +(1-\lambda)(\mathcal{L}_{\text{con}}^u + \mathcal{L}_{\text{cls}}^\mathbf{u}+\mu \mathcal{L}_{\text{reg}})
\end{equation}
where $\lambda$ balances supervised and self-supervised components, and $\mu$ controls regularization strength. This formulation provides the foundation for our methodology addressing both known and unknown $\mathcal{C}^\mathbf{u}$ scenarios.

\subsubsection{Self-attention}
In Vision Transformer (ViT) \cite{ICLR2021ViT}, an input image is divided into a sequence of \texttt{N} fixed-size patches, each treated as a token.
These patch tokens are flattened and concatenated with a $\texttt{[CLS]}$ token to form the input sequence, which is then processed sequentially by $L$ transformer layers.
Let $\mathbf{X}_{l-1}\! \in \! \mathbb{R}^{(1+\texttt{N})\times D}$ denote the input of the $l^\text{th}$ layer, where $D$ is the token dimension.
This input is linearly projected into queries and keys, 
denoted as $\mathbf{Q}_{l-1}$ and $\mathbf{K}_{l-1} \! \in \! \mathbb{R}^{(1+\texttt{N})\times D}$, respectively.
ViT employs a self-attention mechanism to capture dependencies among all patches.
The attention map $\mathcal{A}_{l-1} \! \in \! \mathbb{R}^{(1+\texttt{N})\times (1+\texttt{N})}$ is computed as:
\begin{equation}
\mathcal{A}_{l-1}=\text{softmax}\big(\text{concat}(\mathbf{Q}^{\text{cls}}_{l-1}\mathbf{K}^{\top}_{l-1}, \mathbf{Q}^{\text{patch}}_{l-1}\mathbf{K}^{\top}_{l-1})\big)
\end{equation}
where $\mathbf{Q}_{l-1}$ is divided into a $\texttt{[CLS]}$ token $\mathbf{Q}^{\text{cls}}_{l-1}$ and patch tokens $\mathbf{Q}^{\text{patch}}_{l-1}$.
Multi-head design is omitted for simplicity.

\section{Method}
\label{sec:method}
Fig. \ref{fig:pipeline} illustrates the pipeline of our framework for multi-class anomaly classification.
To extract discriminative anomaly features for each image, 
we introduce Dual-Enhanced Vision Transformer (DE-ViT) built upon the vanilla ViT architecture.
DE-ViT incorporates the Soft-focused Attention that uses an anomaly map to focus on anomalies.
Then, we add Auxiliary Classification Tokens between $\texttt{[CLS]}$ token and patch tokens in the last layer of our DE-ViT. 
These tokens complement the $\texttt{[CLS]}$ token by attending to anomalies it may overlook, thereby enhancing the discriminability between features of similar anomaly classes.
Furthermore, the resulting features are fed into a multi-class classifier, accompanied by an Orthogonal Weight Loss that enlarges the inter-class margin.
For scenarios with an unknown number of anomaly classes, we propose a Correlation-based Number Estimation strategy to estimate the class number during training.
Furthermore, we design the Under-estimation Penalized Score to evaluate the accuracy of the estimated number.

\subsection{Soft-focused Attention}
\label{sec:sf_attn}

\begin{figure}[t]
\begin{center}
\includegraphics[width=0.47\textwidth]{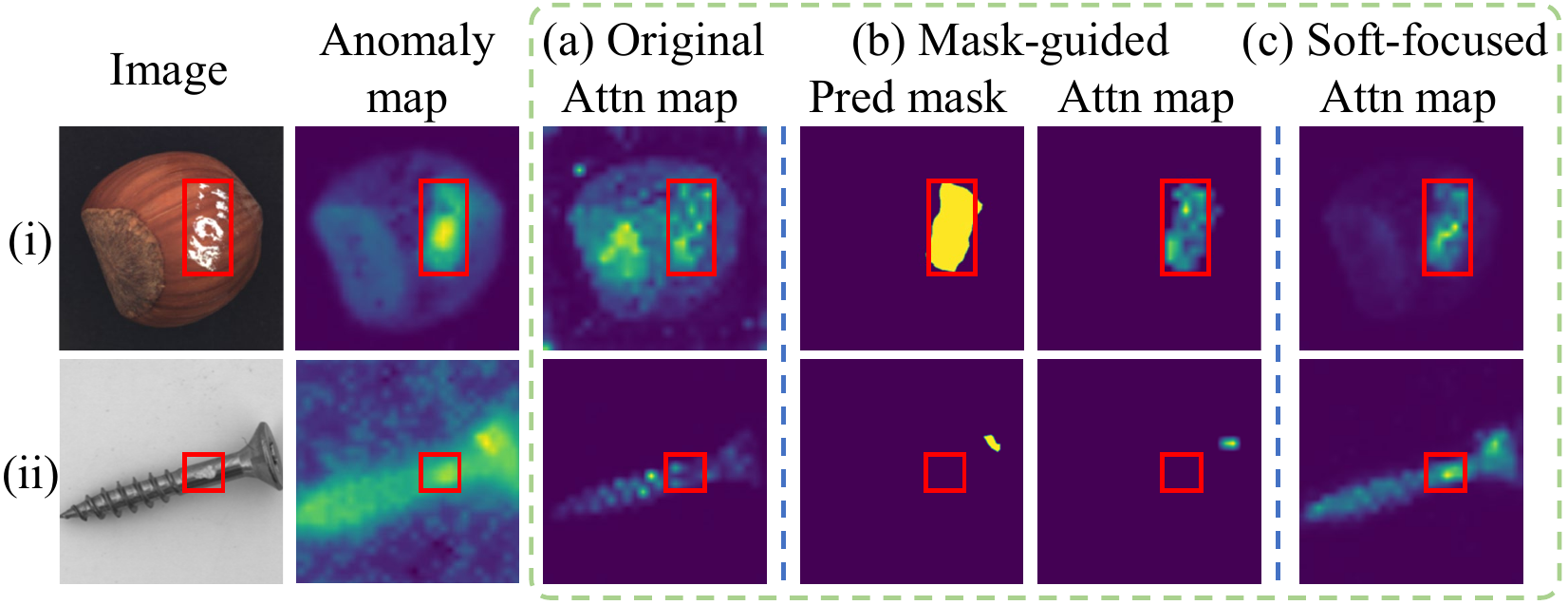}
\vspace{-1.\baselineskip}
\caption{
Attention maps of the $\texttt{[CLS]}$ token of the ViT last layer.
We visualize them using: (a) the original design, (b) mask-guided attention, and (c) soft-focused attention.}
\label{fig:sf_attn}
\end{center}
\end{figure}

In industrial scenarios, anomalies often occupy only small regions within images.
Pre-trained feature extractors like ViT, which are trained on natural images,
struggle to focus on these subtle anomalies.
As shown in Fig. \ref{fig:sf_attn} (a), the $\texttt{[CLS]}$ token attends to the industrial products (i) or background noise (ii) rather than the anomaly itself.
This results in feature representations that conflate anomaly information with product background, confusing the classifier and harming discrimination between different anomaly classes.
To address this, we design a Soft-focused Attention mechanism that uses an anomaly map to guide the model to regions with higher anomaly probabilities,
thereby extracting more discriminative and anomaly-specific features for reliable classification.

Specifically, an anomaly map $A_i \in \mathbb{R}^{H \!\times \!W}$ for image $I_i$ is first obtained using an existing anomaly detection method \cite{iccv2023pni, ICLR2024MuSc, CVPR2022patchcore}.
To assign an anomaly probability to each patch, we downsample $A_i$ to $\sqrt{\texttt{N}}\times\sqrt{\texttt{N}}$ via average pooling and flatten it into a vector of length $\texttt{N}$,
where each element corresponds to a patch's anomaly probability.
To align with the transformer layer input $\mathbf{X}_{l-1}\! \in \! \mathbb{R}^{(1+\texttt{N})\times D}$,
we prepend a constant $1$ to the vector, indicating full attention to the $\texttt{[CLS]}$ token.
We assign the anomaly probabilities of their source patches to these tokens, yielding an anomaly-aware vector $\overline{\mathbf{A}}_{i}\! \in \! \mathbb{R}^{(1+\texttt{N})\times 1}$.
To align the value range with ViT attention, we normalize $\overline{\mathbf{A}}_{i}$ to have the same mean and deviation as the attention scores in the original ViT.
This normalized vector $\hat{\mathbf{A}}_{i}$ then steers the self-attention mechanism toward anomaly regions,
leading to a modified formulation of Eq. (2) as follows:
\begin{equation}
\begin{aligned}
\mathcal{A}_{l\!-\!1}\!=\!\text{softmax}\big(\text{concat}(\mathbf{Q}^{\text{cls}}_{l\!-\!1}\mathbf{K}^{\top}_{l\!-\!1}\!+\!\hat{\mathbf{A}}_{i}, \mathbf{Q}^{\text{patch}}_{l\!-\!1}\mathbf{K}^{\top}_{l\!-\!1})\big)
\end{aligned}
\end{equation}
We apply $\hat{\mathbf{A}}_{i}$ only to the $\texttt{[CLS]}$ token.
In Fig. \ref{fig:sf_attn}, when guided by our soft-focused attention (c), the $\texttt{[CLS]}$ token predominantly attends to regions with higher anomaly probabilities compared to original self-attention (a).
Furthermore, Fig. \ref{fig:sf_attn} (ii) shows that mask-guided attention (b) proposed in AnomalyNCD is sensitive to missed detections (\textcolor[rgb]{1.0,0.0,0.0}{red boxes}), 
due to its reliance on a binary anomaly map.
This binarization discards regions with potential anomalies, 
preventing $\texttt{[CLS]}$ token from attending to undetected anomalies.
In contrast, our soft-focus attention (c) uses the continuous anomaly map to attend to real anomalies, avoiding such hard thresholds.
By focusing on relevant anomalies, the model extracts features that are more specific to anomalies themselves, reducing interference from normal regions and yielding a cleaner, more discriminative representation for classification.

\subsection{Auxiliary Classification Tokens}
\label{sec:cover_token}

\begin{figure}[t]
\begin{center}
\includegraphics[width=0.47\textwidth]{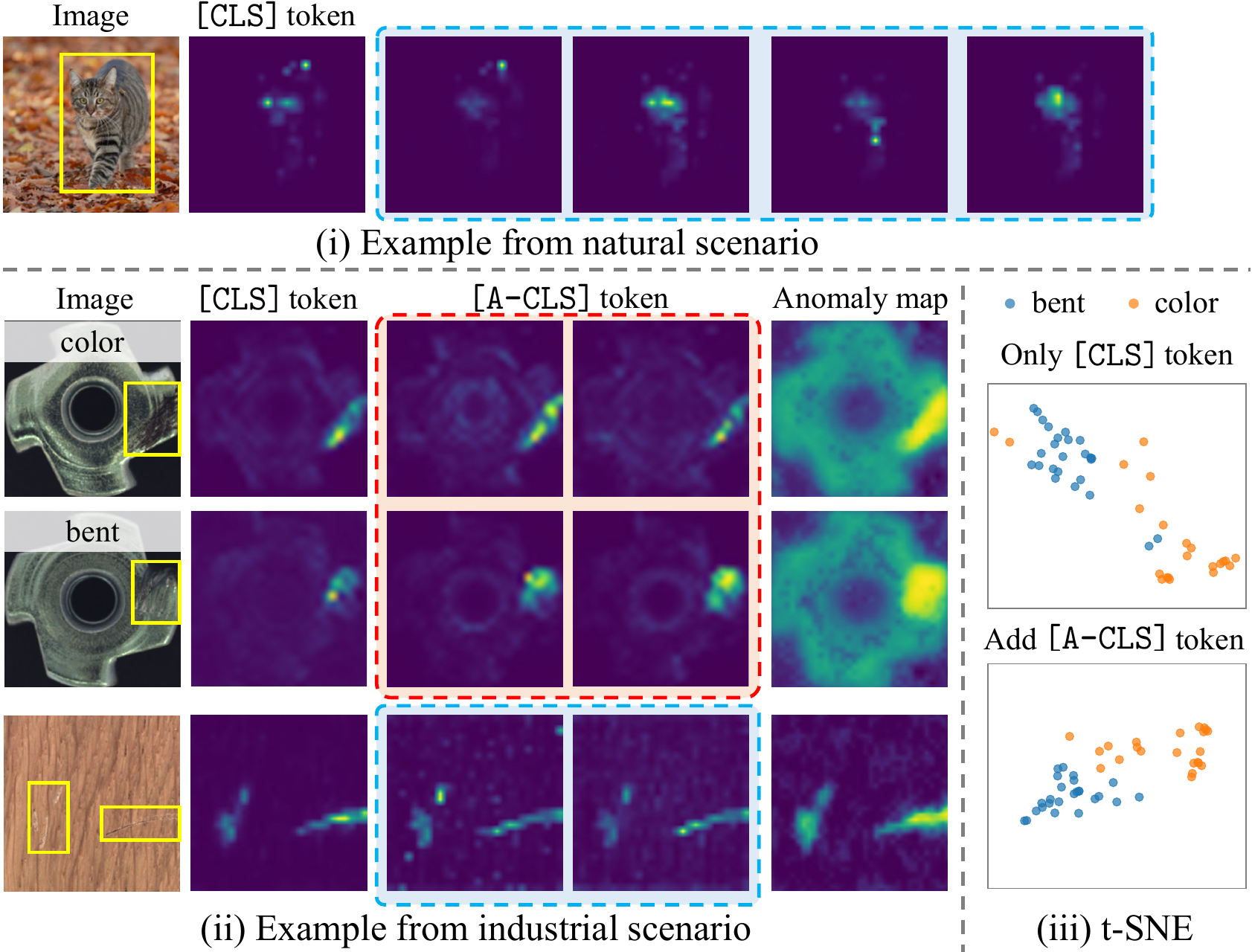}
\vspace{-1.\baselineskip}
\caption{
Attention maps of the $\texttt{[CLS]}$ token, two $\texttt{[ac1]}$ tokens (\textcolor[rgb]{1.0,0.0,0.0}{red boxes}), and two $\texttt{[ac2]}$ tokens (\textcolor[rgb]{0.0,0.69,0.94}{blue boxes}) from the ViT last layer,
illustrated using one natural scene example (i) and two industrial scene examples (ii) and (iii).
}
\label{fig:ca_attn}
\end{center}
\end{figure}

Existing classification methods typically utilize the $\texttt{[CLS]}$ token from ViTs directly for final prediction.
Although our soft-focused attention helps guide this token to abnormal regions,
it may still concentrate on only a local part of an anomaly due to the inherent semantic diversity within anomalies.
As shown in Fig. \ref{fig:ca_attn}, this limited focus brings the incomplete feature representation, which causes features from different anomaly classes to overlap in the embedding space (iii), leading to misclassification.

To address this, we propose Auxiliary Classification \texttt{[A-CLS]} Tokens, which are inserted between $\texttt{[CLS]}$ token and patch tokens in the final transformer layer,
forming as $\mathbf{X}_{l-1}\! \in \! \mathbb{R}^{(1+k+\texttt{N})\times D}$.
\texttt{[A-CLS]} tokens assist the $\texttt{[CLS]}$ token in perceiving various anomaly regions, 
thereby yielding a more holistic and discriminative anomaly representation.
To ensure \texttt{[A-CLS]} tokens can represent anomaly regions like $\texttt{[CLS]}$ token,
we compute the similarity between the $\texttt{[CLS]}$ token and all patch tokens, selecting the top-$k$ most similar ones as \texttt{[A-CLS]} tokens.
Since $\texttt{[CLS]}$ token has already been guided to focus on anomaly regions via soft-focused attention in preceding layers, the patch tokens most similar to it are likely to also correspond to anomaly regions.
The $k$ \texttt{[A-CLS]} tokens are split into two complementary groups: \texttt{[ac1]} and \texttt{[ac2]},
where \texttt{[ac1]} tokens are guided by the anomaly map to focus on high-probability regions, while \texttt{[ac2]} tokens learn autonomously to discover potentially missed detections, both work collaboratively to complement \texttt{[CLS]} token's representation from different scopes.

\textbf{\texttt{[ac1].}}
The first half of the $k$ \texttt{[A-CLS]} tokens are used as \texttt{[ac1]} tokens.
Although \texttt{[CLS]} token is steered by soft-focused attention, it concentrates primarily on the most prominent abnormal region.
\texttt{[ac1]} tokens are introduced to attend to other salient abnormal regions indicated by the anomaly map, complementing the \texttt{[CLS]} token's focus.
To focus on abnormal regions, these tokens are guided by the anomaly map via the soft-focused attention.
Accordingly, Eq. (3) could be rewritten as follows,
\begin{equation}
\begin{aligned}
\mathcal{A}_{l\!-\!1} &\!=\! \text{softmax}\!\big(\text{concat}(\!\mathbf{Q}^{\text{cls}}_{l\!-\!1}\mathbf{K}^{\top}_{l\!-\!1}\!+\!\hat{\mathbf{A}}_{i},\!\mathbf{Q}^{\text{ac1}}_{l\!-\!1}\mathbf{K}^{\top}_{l\!-\!1}\!+\!\hat{\mathbf{A}}_{i},\\
&~~~~~~~~~~~~~~~~~~\mathbf{Q}^{\text{ac2}}_{l\!-\!1}\mathbf{K}^{\top}_{l\!-\!1}, \mathbf{Q}^{\text{patch}}_{l\!-\!1}\mathbf{K}^{\top}_{l\!-\!1})\big)
\end{aligned}
\end{equation}
where $\mathbf{Q}^{\text{ac1}}_{l-1}$ and $\mathbf{Q}^{\text{ac2}}_{l-1}$ denote the queries obtained by linear projection of the \texttt{[ac1]} and \texttt{[ac2]} tokens, respectively.

\textbf{\texttt{[ac2].}}
We notice that anomaly maps are often imperfect and may contain false positives or negatives.
As shown in the third row of Fig. \ref{fig:ca_attn} (ii), the left abnormal region has lower anomaly probability, which causes the \texttt{[CLS]} token to attend almost to the right region.
Such incomplete attention reduces inter-class separability among different anomaly classes and hinders subsequent classification.
Therefore, we retain half of \texttt{[A-CLS]} as \texttt{[ac2]} tokens to autonomously identify anomaly regions without direct guidance.
To ensure the \texttt{[ac2]} tokens focus on different and complementary parts of the anomaly, we apply an orthogonality constraint on their attention patterns.
Specifically, we compute the attention matrix $\mathcal{A}_{l-1}^\text{ac2}\!\in\!\mathbb{R}^{\frac{k}{2} \times \texttt{N}}$ between the $\frac{k}{2}$ \texttt{[ac2]} tokens (as queries) and all $\texttt{N}$ patch tokens (as keys).
After normalization, $\overline{\mathcal{A}}_{l-1}^\text{ac2}=\mathcal{A}_{l-1}^\text{ac2} / \Vert\mathcal{A}_{l-1}^\text{ac2}\Vert_2$, we define the cosine similarity-based orthogonality loss as:
\begin{equation}
\mathcal{L}_{ac}=\sum_{i \neq j}\overline{\mathcal{A}}_{l-1}^\text{ac2}(i) \overline{\mathcal{A}}_{l-1}^\text{ac2}(j)^\top
\end{equation}
This loss minimizes the similarity between attention vectors of different \texttt{[ac2]} tokens, encouraging each token to focus on a unique region of the anomaly.

As shown in row 1-2 of Fig. \ref{fig:ca_attn} (ii), the \texttt{[ac1]} tokens in \textcolor[rgb]{1.0,0.0,0.0}{red boxes}, guided by soft-focused attention, attend to high-probability regions within the anomaly map, complementing regions overlooked by \texttt{[CLS]} token.
In row 3, however, the anomaly map assigns higher probability only to the right-side anomaly, causing the \texttt{[CLS]} token to miss the left-side region.
The unguided \texttt{[ac2]} tokens (\textcolor[rgb]{0.0,0.69,0.94}{blue boxes}) are not constrained by the anomaly map, and successfully attend to the low-probability, missed-detection region on the left.
This demonstrates their ability to adaptively discover missed anomalies during training.
The t-SNE visualization in (iii) further confirms that the \texttt{[A-CLS]} tokens complement the \texttt{[CLS]} token, yielding a more complete anomaly representation, that reduces inter-class confusion and improves separation.
A similar phenomenon is also observed in natural images, as shown in Fig. \ref{fig:ca_attn} (i), the \texttt{[ac2]} tokens attend to regions distinct from those focused on by the \texttt{[CLS]} token, but those consistently highlight regions in the cat.
These tokens adaptively seek out other discriminative regions beneficial for classification during training, thereby complementing the representation captured by \texttt{[CLS]} token.

\textbf{Token Fusion.}
To extract a discriminative anomaly feature for each image, 
the output from the final layer of our proposed DE-ViT, denoted as $\mathbf{X}_{L}\!\in\!\mathbb{R}^{(1+k+\texttt{N}) \times D}$,
is partitioned into three components: $\texttt{[CLS]}$ token $\mathbf{X}_{L}^{\text{cls}}  \!\in\!\mathbb{R}^{1\times D}$, \texttt{[A-CLS]} tokens $\mathbf{X}_{L}^{\text{a-cls}} \!\in\!\mathbb{R}^{k\times D}$, and patch tokens $\mathbf{X}_{L}^{\text{patch}} \!\in\!\mathbb{R}^{\texttt{N}\times D}$.
The $k$ \texttt{[A-CLS]} tokens are averaged to yield a feature $\mathbf{F}^{\text{a-cls}} \!\in\!\mathbb{R}^{1\times D}$.
For patch tokens, we follow \cite{WACV2023AC} to aggregate them using the anomaly map, to derive the feature $\mathbf{F}^{\text{patch}}  \!\in\!\mathbb{R}^{1\times D}$.
The final multi-class anomaly feature $\overline{\mathbf{F}}\!\in\!\mathbb{R}^{1 \times D}$ obtained by averaging $\mathbf{X}_{L}^{\text{cls}}$, $\mathbf{F}^{\text{a-cls}}$ and $\mathbf{F}^{\text{patch}}$.

\subsection{Orthogonal Weight Loss}
\label{sec:ortho}
The anomaly feature $\overline{\mathbf{F}}$ extracted by DE-ViT is fed into a linear classifier $\mathcal{H}(\cdot)$ to obtain the class prediction.
The classifier's weight matrix $\mathbf{W}\!\in\!\mathbb{R}^{(\mathcal{C}^\textbf{l}+\mathcal{C}^\textbf{u}) \times D}$ represents the prototypes for the $\mathcal{C}^\textbf{l}+\mathcal{C}^\textbf{u}$ labeled and unlabeled classes.
To ensure these prototypes remain well-separated in the feature space, we impose an orthogonality constraint via the following loss:
\begin{equation}
\mathcal{L}_{ow}=\text{sum}(\Vert \mathbf{W}\mathbf{W}^\top - \mathbf{I} \Vert_F^2)
\end{equation}
where $\mathbf{I}$ is the identity matrix and $\Vert \cdot \Vert_F$ denotes the Frobenius norm.
This loss operates on the classifier parameters, encouraging inter-class separation from a prototype perspective. 
The final training loss of our method is calculated as,
\begin{equation}
\mathcal{L}_{total}=\mathcal{L}_\text{NCD} + \mathcal{L}_{ac} + \mathcal{L}_{ow}
\end{equation}

\begin{table*}[tb]
  \centering
  \resizebox{1.0\linewidth}{!}{
  \begin{tabular}{cccccccccccc}
    \toprule
    \multirow{2}{*}{Datasets} & \multirow{2}{*}{Metric} & \multirow{2}{*}{\textbf{IIC}} & \multirow{2}{*}{\textbf{GATCluster}} & \multirow{2}{*}{\textbf{SCAN}} & \multirow{2}{*}{\textbf{UNO}} & \multirow{2}{*}{\textbf{GCD}} & \multirow{2}{*}{\textbf{SimGCD}} &  \multirow{2}{*}{\textbf{AMEND}} &  \textbf{AC} & \textbf{MuSc} & \textbf{MuSc} \\
          &  &  &  & & &  &   &  & \textbf{(Unsup.)} & \textbf{+AnomalyNCD} & \textbf{+MACO (Ours)} \\
    \midrule
    \multirow{2}{*}{MVTec AD}  & NMI & 0.093 & 0.136 & 0.210 & 0.146  & 0.417 & 0.452 & 0.431 & 0.525 & 0.613 & \textbf{0.657} \\
    \multirow{2}{*}{\cite{CVPR2019mvtec}} & ARI & 0.020 & 0.053 & 0.103 & 0.052  & 0.302  &  0.346 & 0.333 & 0.431 & 0.526 & \textbf{0.591} \\
    & $F_1$ & 0.285 & 0.264 & 0.335 & 0.342 & 0.553 & 0.569 & 0.542 & 0.604 & 0.712 & \textbf{0.749} \\
    \midrule
    \multirow{2}{*}{MTD} & NMI & 0.064 & 0.028 & 0.041 & 0.034  & 0.211 & 0.105 & 0.138 & 0.179 & 0.268 & \textbf{0.422} \\
    \multirow{2}{*}{\cite{MTD2020surface}} & ARI & 0.020 & 0.009 & 0.029 & 0.011 & 0.115 & 0.048 & 0.067 & 0.120 & 0.228 & \textbf{0.391} \\
    & $F_1$ & 0.252 & 0.243 & 0.282 & 0.221 & 0.381 & 0.293 & 0.324 & 0.346 & 0.509 & \textbf{0.631} \\
  \bottomrule
  \end{tabular}
  }
  \caption{Quantitative results on the MVTec AD and MTD datasets if the unlabeled class number $\mathcal{C}^\mathbf{u}$ is \textbf{known}.}
  \label{tab:main_known}
\end{table*}

\subsection{Correlation-based Number Estimation}
\label{sec:num_est}
Existing multi-class anomaly classification methods \cite{pr2026uniformaly,cvpr2025anomalyncd,cvpr2024blindlca} assume the number of unlabeled anomaly classes $\mathcal{C}^\textbf{u}$ is known, this quantity is often unavailable in complex industrial environments.
To address this, we propose a Correlation-based Number Estimation strategy, which estimates $\mathcal{C}^\textbf{u}$ by leveraging the labeled classes with known number $\mathcal{C}^\textbf{l}$.

\begin{figure}[t]
\begin{center}
\includegraphics[width=0.47\textwidth]{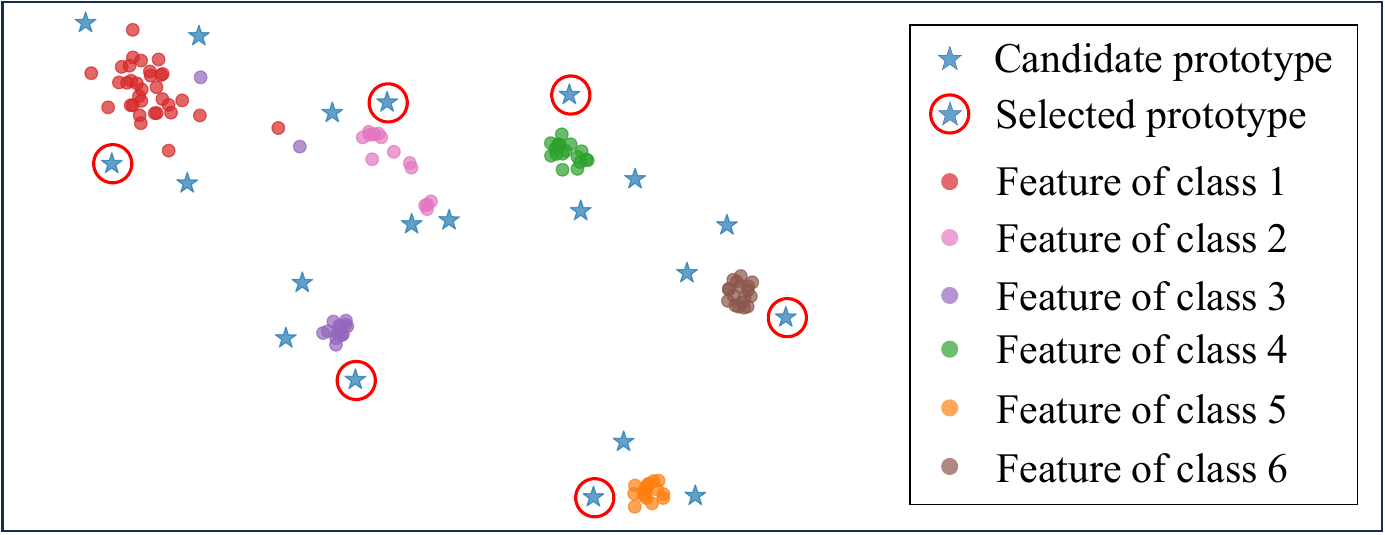}
\vspace{-1.\baselineskip}
\caption{
t-SNE visualization of anomaly features $\overline{\mathbf{F}}$ with 6 classes and classifier prototypes $\mathbf{W}^\textbf{u}$, where stars are all prototypes and circled stars denote selected ones.
}
\label{fig:est_tsne}
\vspace{-0.8\baselineskip}
\end{center}
\end{figure}

Before training, we initialize the classifier with a predefined upper bound $\mathcal{C}_\text{max}$ as the initial number of unlabeled classes $\mathcal{C}^\textbf{u}$.
This provides a safe over-estimation to ensure all potential classes are covered.
Our model is first trained for $e$ epochs using $\mathcal{C}_\text{max}$ as $\mathcal{C}^\textbf{u}$.
At epoch $e$, our strategy estimates the refined number $\hat{\mathcal{C}}^\textbf{u}$, and the model is fine-tuned for the remaining epochs using this estimated $\hat{\mathcal{C}}^\textbf{u}$.

When estimating the number of classes, the classifier weights $\mathbf{W}$ are split into labeled and unlabeled portions as $\mathbf{W}^\textbf{l}\!\in\!\mathbb{R}^{\mathcal{C}^\textbf{l} \times D}$ and $\mathbf{W}^\textbf{u}\!\in\!\mathbb{R}^{\mathcal{C}_\text{max}\times D}$.
Then we compute the correlation matrices $\mathcal{S}^\textbf{l}=\mathbf{W}^\textbf{l}\cdot{\mathbf{W}^\textbf{l}}^{\top}$ and $\mathcal{S}^\textbf{u}=\mathbf{W}^\textbf{u}\cdot{\mathbf{W}^\textbf{u}}^{\top}$ to capture pairwise similarities among class prototypes $\mathbf{W}^\textbf{l}$ and $\mathbf{W}^\textbf{u}$, respectively.
The average value of the entries in $\mathcal{S}^\textbf{l}$ defines a threshold $\overline{\lambda}$, reflecting the distance among prototypes of labeled classes.
We then use $\overline{\lambda}$ as a reference to assess whether unlabeled prototypes are sufficiently distinct, thereby transferring separability cues from labeled to unlabeled classes.
Our strategy starts by randomly choosing one prototype from $\mathbf{W}^\textbf{u}$ as the initial seed.
At each subsequent step, we compute for every remaining candidate prototype its maximum similarity to any prototype already selected, based on $\mathcal{S}^\textbf{u}$.
The candidate with the smallest such maximum similarity is then added to the selected set, until its value becomes greater than $\overline{\lambda}$.
This process dynamically reduces the initial over-estimated $\mathcal{C}_\text{max}$ to a more accurate estimate $\hat{\mathcal{C}}^\textbf{u}$.
As shown in Fig. \ref{fig:est_tsne}, multiple candidate prototypes reside around each cluster.
Our strategy selects the most representative prototype for each cluster, ensuring that every cluster is represented by one prototype.
This avoids splitting the same class or merging different classes.
See Appendix for more details.

\subsection{Under-estimating Penalized Score}
\label{sec:ups}
When the number of unlabeled anomaly classes is unknown, the Mean Absolute Percentage Error (MAPE) is commonly used to evaluate the accuracy of the estimated number $\hat{\mathcal{C}}^\textbf{u}$.
However, MAPE treats underestimation and overestimation equally, which is unsuitable for industrial settings.
Underestimation (i.e., $\hat{\mathcal{C}}^\textbf{u}\!<\!\mathcal{C}^\textbf{u}$) is critically harmful, as it causes multiple distinct anomaly types to be merged into a single class, obscuring fault patterns, hindering root‑cause analysis, and targeted process improvements.
In contrast, overestimation ($\hat{\mathcal{C}}^\textbf{u}\!>\!\mathcal{C}^\textbf{u}$) primarily results in a manageable increase in manual verification effort, a less severe cost compared to missing anomalies.
To address this bias, we propose the Under-estimation Penalized Score (UPS), which assigns asymmetric penalties to underestimation and overestimation.

We define three desired properties for the evaluation metric:
\large{\textcircled{\scriptsize{1}}}\normalsize The score equals $1$ when $\hat{\mathcal{C}}^\textbf{u}\!=\!\mathcal{C}^\textbf{u}$, indicating perfect estimation accuracy.
\large{\textcircled{\scriptsize{2}}}\normalsize It increases monotonically when $\hat{\mathcal{C}}^\textbf{u}\!<\!\mathcal{C}^\textbf{u}$, reaching $0$ at $\hat{\mathcal{C}}^\textbf{u}\!=\!1$ (since at least ``normal'' and ``anomaly'' classes must exist).
\large{\textcircled{\scriptsize{3}}}\normalsize It decreases monotonically when $\hat{\mathcal{C}}^\textbf{u}\!>\!\mathcal{C}^\textbf{u}$, asymptotically approaching 0.
Based on these properties, we design our UPS as follows:
\begin{equation}
    \text{UPS}(\hat{\mathcal{C}}^\textbf{u},\mathcal{C}^\textbf{u}) =
    \left \{ 
    \begin{array}{ll}
    \frac{\hat{\mathcal{C}}^\textbf{u}-1}{\mathcal{C}^\textbf{u}-1},~~\text{if}~~ \hat{\mathcal{C}}^\textbf{u}\!<\!\mathcal{C}^\textbf{u} \\[1mm]
    \frac{\mathcal{C}^\textbf{u}}{\hat{\mathcal{C}}^\textbf{u}},~~~~~\text{if}~~\hat{\mathcal{C}}^\textbf{u}\!\geq\!\mathcal{C}^\textbf{u},
    \end{array}\right.
\end{equation}

At $\hat{\mathcal{C}}^\textbf{u}\!=\!\mathcal{C}^\textbf{u}$, the slope of $\frac{\hat{\mathcal{C}}^\textbf{u}-1}{\mathcal{C}^\textbf{u}-1}$ is $\frac{1}{\mathcal{C}^\textbf{u}-1}$, while the magnitude of the slope for $\frac{\mathcal{C}^\textbf{u}}{\hat{\mathcal{C}}^\textbf{u}}$ is $\frac{1}{\mathcal{C}^\textbf{u}}$.
This ensures a steeper penalty for underestimation ($\hat{\mathcal{C}}^\textbf{u}\!<\!\mathcal{C}^\textbf{u}$) than for overestimation.

\section{Experiments}
\subsection{Experimental Setting}
\subsubsection{Datasets}
Our experiments use industrial datasets MVTec AD \cite{CVPR2019mvtec} and MTD \cite{MTD2020surface}.
The MVTec AD includes images across 10 object categories and 5 texture categories.
Each category contains a minimum of two anomaly types.
To ensure a fair evaluation, we follow the setup in \cite{WACV2023AC, cvpr2025anomalyncd} to exclude the combined anomaly classes.
The MTD comprises 952 normal and 392 abnormal images, with the anomalies categorized into five classes.
We adopt the protocol from \cite{wacv2021differnet}, using only 20\% of normal images for testing.
Following \cite{cvpr2025anomalyncd}, the labeled set $\mathcal{D}^\mathbf{l}$ is derived from the Aero-engine Blade Anomaly Detection Dataset (AeBAD-S) \cite{AeBAD}.

\subsubsection{Evaluation protocol}
To evaluate the performance, we use two clustering metrics (normalized mutual information (NMI) \cite{NMI}, and adjusted rand index (ARI) \cite{ARI}) and one classification metric ($F_1$ score).
The Hungarian algorithm \cite{hungarian} is employed to align predicted clusters with ground-truth labels.
For scenarios with an unknown class number, we assess estimation accuracy using the Mean Absolute Percentage Error (MAPE) \cite{IJF2006MAPE} and our proposed UPS.

\subsubsection{Implementation details}
Following \cite{cvpr2025anomalyncd}, we adopt a ViT-B/8 model pretrained with DINO \cite{iccv2021dino} as our feature extractor.
The input images are resized to $224 \times 224$ resolution.
In the Soft-focused Attention, we use zero-shot AD method MuSc \cite{ICLR2024MuSc} to generate anomaly maps by default.
We also provide more results using other AD methods in the Appendix.
The number of Auxiliary Classification Tokens is set to 8.
We perform our Correlation-based Number Estimation at epoch $e=10$ and the predefined class number $\mathcal{C}_\text{max}$ is set to 20.
We conduct experiments across all datasets according to the same configuration.

\subsection{Comparison with the State-of-the-Arts}
\subsubsection{Unlabeled classes number $\mathcal{C}^\mathbf{u}$ is known.}
We compare our approach with two state-of-the-art multi-class anomaly classification methods: AC \cite{WACV2023AC} and AnomalyNCD \cite{cvpr2025anomalyncd}.
We also include some natural-scene NCD methods: ICC \cite{ICCV2019IIC}, GATCluster \cite{niu2020gatcluster}, SCAN \cite{van2020scan}, UNO \cite{ICCV2021UNO}, GCD \cite{CVPR2022GCD}, SimGCD \cite{ICCV2023SimGCD}, and AMEND \cite{WACV2024amend}.
All methods use the AeBAD-S dataset as labeled set.
Table \ref{tab:main_known} summarizes the multi-class anomaly classification results on MVTec AD \cite{CVPR2019mvtec} and MTD \cite{MTD2020surface} datasets.
These natural-scene NCD methods often struggle to localize small anomalies, leading to generally lower scores.
Our method outperforms both industrial-specific approaches across all metrics, improving ARI by 6.5\% on MVTec AD.
On the more challenging MTD dataset, which contains noisier anomaly maps, our soft-focus attention effectively mitigates the impact of missed detections, yielding gains of 16.3\% in ARI.

\begin{table}[tb]
  \centering
  \resizebox{1.\linewidth}{!}{
  \begin{tabular}{llccccccc}
    \toprule
    Datasets & Metric & ~~w/o~~ & DTC & DAEM & GCD & CMS & ProtoGCD & \textbf{Ours} \\
    \midrule
    \multirow{2}{*}{MVTec AD} & MAPE $\downarrow$ & 1.450 & 0.490 & 0.450 & 0.560 & 0.610 & 0.600 & \textbf{0.384} \\
    & UPS $\uparrow$ & 0.491 & 0.576 & 0.517 & 0.323 & 0.267 & 0.563 & \textbf{0.817} \\
    \midrule
    \multirow{2}{*}{MTD} & MAPE $\downarrow$ & 0.830 & 0.500 & 0.500 & 0.670 & 0.670 & 0.330 & \textbf{0.000} \\
    & UPS $\uparrow$ & 0.546 & 0.400 & 0.400 & 0.200 & 0.200 & 0.600 & \textbf{1.000} \\
  \bottomrule
  \end{tabular}
  }
  \caption{Compared with different number estimate strategies on two industrial datasets.  Best results in bold.}
  \label{tab:abl_est}
\end{table}

\subsubsection{Unlabeled classes number $\mathcal{C}^\mathbf{u}$ is unknown.}
We compare Correlation-based Number Estimation strategy with other class number estimating methods in Table \ref{tab:abl_est} and achieve the best results on both MAPE and UPS metrics.
On MVTec AD, we improve UPS by 24.1\% over the best prior method.
On MTD, we attain perfect prediction (MAPE=0.000, UPS=1.000).
Most current methods indirectly estimate the number of unlabeled classes by evaluating clustering performance on labeled classes.
However, since labeled classes are trained with supervised loss, their features become compact more rapidly, causing clustering metrics to saturate near 100\% early in training.
This makes it difficult to reliably estimate the number of unlabeled classes based on labeled-class clustering results.

\subsection{Ablation Study}

\begin{table}[t]
  \centering
  \resizebox{1.\linewidth}{!}{
   \begin{tabular}{lcccccc}
    \toprule
    & \multicolumn{3}{c}{MVTec AD} & \multicolumn{3}{c}{MTD} \\
    \cmidrule(l){2-4} \cmidrule(l){5-7} 
    Attention & ~NMI~ & ~ARI~ & ~$F_1$~ & ~NMI~ & ~ARI~ & ~$F_1$~ \\
    \midrule
    (a) w/o guidance & 0.587 & 0.522 & 0.689 & 0.305 & 0.217 & 0.485 \\
    (b) mask-guided & 0.647 & 0.565 & 0.739 & 0.254 & 0.208 & 0.463 \\
    (c) soft-focus (Ours) & \textbf{0.657} & \textbf{0.591} & \textbf{0.749} & \textbf{0.422} & \textbf{0.391} & \textbf{0.631} \\
  \bottomrule
  \end{tabular}
  }
  \caption{Compared with different attention mechanisms.}
   \label{tab:abl_sfattn}
\end{table}

\begin{table}[t]
  \centering
  \resizebox{1.\linewidth}{!}{
   \begin{tabular}{lcccccc}
    \toprule
    & \multicolumn{3}{c}{MVTec AD} & \multicolumn{3}{c}{MTD} \\
    \cmidrule(l){2-4} \cmidrule(l){5-7} 
    Attention & ~NMI~ & ~ARI~ & ~$F_1$~ & ~NMI~ & ~ARI~ & ~$F_1$~ \\
    \midrule
    (a) w/o all $\texttt{[A-CLS]}$ tokens & 0.651 & 0.581 & 0.748 & 0.390 & 0.335 & 0.588 \\
    (b) w $\texttt{[ac1]}$ tokens & 0.641 & 0.578 & 0.730 & 0.408 & 0.350 & 0.588 \\
    (c) w $\texttt{[ac2]}$ tokens & 0.632 & 0.569 & 0.726 & \textbf{0.428} & \textbf{0.403} & 0.629 \\
    (d) Ours & \textbf{0.657} & \textbf{0.591} & \textbf{0.749} & 0.422 & 0.391 & \textbf{0.631} \\
  \bottomrule
  \end{tabular}
  }
  \caption{Ablation studies on auxiliary classification tokens.}
   \label{tab:abl_catoken}
\end{table}

\subsubsection{Discussion of the soft-focus attention.}
Table \ref{tab:abl_sfattn} ablates our soft-focus attention.
Compared to using no guidance (a), our approach focuses on regions with higher anomaly probabilities via the anomaly map and achieves better performance.
Against the mask-guided attention (b) from AnomalyNCD, our method reduces the propagation of cumulative errors from the binarization stage into the downstream multi-class framework.
\textbf{This is critical, as post-binarization missed detection rates reach 16.02\% on MVTec AD and 40.46\% on MTD.}
Consequently, our method increases ARI by 2.6\% on MVTec AD.
On MTD, where predicted masks are less accurate, a more substantial ARI gain of 18.3\% is observed.
Further visualizations on MTD are provided in the Appendix.

\subsubsection{Effectiveness of \texttt{[A-CLS]} tokens.}
We conduct ablations of the auxiliary classification (\texttt{[A-CLS]}) tokens in Table \ref{tab:abl_catoken}, comparing three settings: (a) without any \texttt{[A-CLS]} tokens, (b) using only the anomaly map-guided \texttt{[ac1]} tokens, and (c) using only the unguided \texttt{[ac2]} tokens.
Employing both \texttt{[ac1]} and \texttt{[ac2]} tokens yields the best overall performance across most metrics.
However, on some metrics, using only the \texttt{[ac2]} tokens (c) leads to a slight improvement.
This occurs because, when the AD methods introduce significant over-detection or under-detection (particularly on the MTD dataset), the anomaly map may provide misleading guidance, causing \texttt{[ac1]} tokens to focus on incorrect regions.
These \texttt{[A-CLS]} tokens could also be applied to other visual and non-visual tasks reported in the Appendix.

\begin{table}[t]
    \centering
    \setlength{\tabcolsep}{2.5mm}
    \resizebox{1.\linewidth}{!}{
    \begin{tabular}{cccccc}
        \toprule
        & \multicolumn{2}{c}{\texttt{[CLS]} token} & \multicolumn{3}{c}{+\texttt{[A-CLS]} tokens} \\ 
        \cmidrule(l){2-3} \cmidrule(l){4-6} 
         & w/o sf-attn & w sf-attn & only \texttt{[ca1]} & only \texttt{[ca2]} & \texttt{[ca1]}+\texttt{[ca1]} \\  
        \midrule
        TP regions & 22.05\% & 35.02\% & 39.26\% & 40.55\% & 42.45\% \\ 
        FN regions & 9.86\% & 10.68\% & 12.34\% & 20.79\% & 21.37\% \\
        \bottomrule
    \end{tabular}
    }
    \caption{Quantification of anomaly attention alignment on all abnormal images of MVTec AD and MTD datasets.}
    \label{tab:align_quantitative}
\end{table}

\subsubsection{Discussion of anomaly attention alignment.}
Both soft-focus attention (sf-attn) and auxiliary classification tokens (\texttt{[A-CLS]}) are used to facilitate the alignment of attention to actual abnormal regions.
To quantitatively evaluate the attention alignment, we statistically measure the attention focus on true‑positive (TP) and false‑negative (FN) regions across all abnormal samples in MVTec AD and MTD. Results in Table \ref{tab:align_quantitative} show that soft‑focused attention drives \texttt{[CLS]} token to attend more to TP regions. The guided \texttt{[ca1]} tokens complement this by focusing on other TP regions, while unguided \texttt{[ca2]} tokens attend more to FN regions. This confirms that our mechanism improves attention coverage of both visible and missed anomalies, explaining the performance gain.

\begin{table}[h]
  \centering
  \resizebox{0.9\linewidth}{!}{
   \begin{tabular}{lcccccc}
    \toprule
    & \multicolumn{3}{c}{MVTec AD} & \multicolumn{3}{c}{MTD} \\
    \cmidrule(l){2-4} \cmidrule(l){5-7} 
    Loss & ~NMI~ & ~ARI~ & ~$F_1$~ & ~NMI~ & ~ARI~ & ~$F_1$~ \\
    \midrule
    w/o $\mathcal{L}_{ac}$ & 0.656 & 0.588 & 0.748 & 0.419 & 0.387 & 0.626 \\
    w/o $\mathcal{L}_{ow}$ & \textbf{0.659} & 0.586 & \textbf{0.749} & 0.416 & 0.380 & 0.626 \\
    Ours & 0.657 & \textbf{0.591} & \textbf{0.749} & \textbf{0.422} & \textbf{0.391} & \textbf{0.631} \\
  \bottomrule
  \end{tabular}
  }
  \caption{Ablations on the composition of the loss function.}
  \label{tab:abl_loss}
\end{table}

\subsubsection{Effect of $\mathcal{L}_{ac}$ and $\mathcal{L}_{ow}$.}
Our method extends the traditional NCD loss with two terms: $\mathcal{L}_{ac}$ and $\mathcal{L}_{ow}$.
As shown in Table \ref{tab:abl_loss}, $\mathcal{L}_{ac}$ encourages separation among different auxiliary classification tokens, improving the $F_1$ by 0.5\% on the MTD dataset.
The orthogonal weight loss $\mathcal{L}_{ow}$ applied to the classifier increases ARI by 1.1\% on the same dataset.
However, it leads to a slight NMI decrease of 0.2\%, as forcing prototype separation for visually similar anomaly classes may reduce clustering flexibility when features overlap significantly.

\section{Conclusion}
This paper presents MACO for multi-class anomaly classification, tackling three challenges: noisy, incomplete anomaly representation, and unknown class number.
We propose a soft-focus attention to focus on anomaly regions and Auxiliary Classification Tokens to capture diverse anomaly patterns.
The Correlation-based Number Estimation strategy infers class number with lower overhead.
Experiments on MVTec AD and MTD show the state-of-the-art performance under both known and unknown class number settings.

\bibliography{aaai2027}

\end{document}